\def\year{2020}\relax
\documentclass[letterpaper]{article}
\usepackage{aaai20}
\usepackage{times}
\usepackage{helvet}
\usepackage{courier}
\usepackage[hyphens]{url} 
\usepackage{graphicx}
\usepackage{graphicx}
\usepackage{array}
\usepackage{balance}
\usepackage{subfigure}
\usepackage{multirow}
\usepackage{float}
\usepackage{color}
\usepackage{soul}
\usepackage{mdframed}
\usepackage{amsopn}
\usepackage{mathrsfs}
\usepackage{mathtools}
\usepackage{amsmath}
\usepackage{amssymb}
\usepackage{bbm}
\usepackage{arydshln}
\usepackage{multicol}
\usepackage{blkarray}
\usepackage{enumerate}
\usepackage{courier}
\usepackage{rotating}
\usepackage{booktabs}
\usepackage{diagbox}
\usepackage{fancybox}
\usepackage{minibox}
\usepackage{cases}
\usepackage[noend]{algpseudocode}
\usepackage{algorithm}
\newcommand{\newcite}[1]{\citeauthor{#1} \shortcite{#1}}

\newcommand{\ourmethod}{ABSent}
\newcommand{\ourmethoduni}{uni-Sent}

\title{\ourmethod: Cross-Lingual Sentence Representation Mapping \\with Bidirectional GANs}

\author{Zuohui Fu,  \textsuperscript{\rm}
Yikun Xian,  \textsuperscript{\rm}
Shijie Geng,  \textsuperscript{\rm}
Yingqiang Ge,  \textsuperscript{\rm}\\
\bf \Large Yuting Wang,  \textsuperscript{\rm}
Xin Dong,  \textsuperscript{\rm}
Guang Wang,  \textsuperscript{\rm}
Gerard de Melo\textsuperscript{\rm}\\
\textsuperscript{\rm}Department of Computer Science\\ Rutgers University, New Brunswick, NJ, USA\\
zuohui.fu@rutgers.edu, siriusxyk@gmail.com, \{sg1309, yingqiang.ge\}@rutgers.edu, \\
yw632@cs.rutgers.edu, xd48@rutgers.edu, gw255@cs.rutgers.edu, gdm@demelo.org
}

\begin{document}

\maketitle

\begin{abstract}
A number of cross-lingual transfer learning approaches based on neural networks have been proposed for the case when large amounts of parallel text are at our disposal. However, in many real-world settings, the size of parallel annotated training data is restricted. Additionally, prior cross-lingual mapping research has mainly focused on the word level. 
This
raises the question of whether such techniques can also be applied
to effortlessly obtain cross-lingually aligned sentence representations. 
To this end, we propose an Adversarial Bi-directional Sentence Embedding Mapping (\ourmethod) framework, which learns mappings of cross-lingual sentence representations from limited quantities of parallel data.
The experiments show that our method outperforms several technically more powerful approaches, especially under challenging low-resource circumstances.
The source code is available from \url{https://github.com/zuohuif/ABSent} along with relevant datasets.
\end{abstract}

\section{Introduction}

Not only have regular vector representations of words become ubiquitous \cite{word2vec}, but bilingual word embeddings \cite{DBLP:journals/corr/Ruder17} have as well enjoyed remarkable success, enabling many novel forms of cross-lingual NLP. Owing to the availability of adequate cross-lingual word datasets and supervised learning methods, cross-lingual transfer learning at the word level is  well-studied. However, only few results exist on sentence-level cross-lingual mapping, let alone studying low-resource settings.

In order to effortlessly obtain sentence-level representations, one of the most popular methods is to compute a (possibly weighted) average of the word vectors of all words encountered in a sentence. The method is favored for its straightforward simplicity, particularly in light of the widespread availability of pre-trained word vectors. While there 
are numerous more powerful methods (cf.\ Section \ref{sec:related-work}), 
they require substantially longer training times and pre-trained models are typically less convenient to load. 
Somewhat surprisingly, weighted averages of word vectors
have been shown capable of outperforming several more advanced techniques, including certain Long short-term memory (LSTM)-based setups
\cite{wieting2015towards,arora2017asimple}. 
\begin{figure}[t]
\centering
\includegraphics[width=.95\columnwidth]{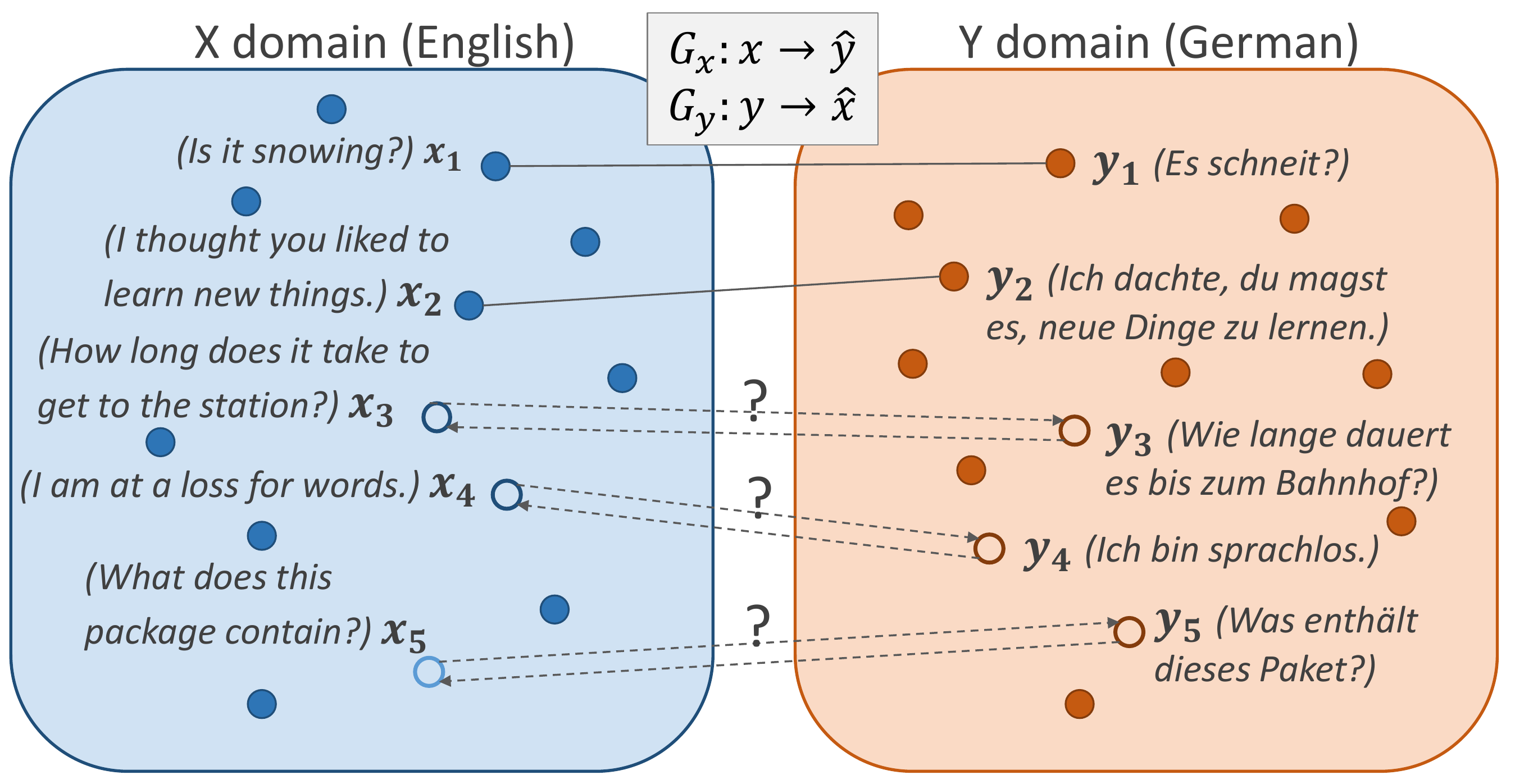}
\caption{Sentence alignment between the respective embeddings of English and German sentences. Solid circles represent training samples, while hollow circles represent test samples. The challenge is to align test samples (connected by dotted segments) with limited numbers of training pairs (connected by solid segments) and a large number of unpaired training samples.}
\label{fig:motivation} 
\end{figure}
In this paper, we study to what extent sentence embeddings based on simple word vector averages can be aligned cross-lingually.
While word vector averages have been studied for embeddings of entire text documents, such document embeddings mainly need to capture topic information. Sentence embeddings, in contrast, are typically expected to retain more detailed semantic information. If simple word vector averages can achieve this cross-lingually despite being entirely oblivious of the order of words in the input sentences, this would provide a simple means of connecting semantically related sentences across language boundaries, in support of a diverse range of possible tasks such as question answering, recommendation, plagiarism detection \cite{liu2019xqa,Xian:2019:RKG:3331184.3331203,ferrero2017usingword}.
While there has been research on joint multilingual training of NMT to obtain richer cross-lingual sentence embeddings \cite{Wang2019MultilingualNM}, such methods tend to require substantial training data.

At the same time, aligning sentences cross-lingually with limited parallel data is challenging, as it is not obvious how to exploit non-parallel data
(see Figure \ref{fig:motivation}).
While linear transformations have proven fruitful for cross-lingual word vector mapping \cite{Mikolov2013ExploitingSA}, recent work shows that non-linear transformations may be necessary even just at the level of individual words \cite{Nakashole2018}. At the sentence level, this necessity may be much more pronounced, due to the divergent syntactic and morphological properties of different languages and the linear superposition of different concepts. We not only study this empirically but also show how non-linear transformations can be learned with limited parallel data.

Specifically, we propose Adversarial Bi-directional Sentence Embedding Mapping (\ourmethod), based on the generative adversarial network (GAN) framework to bridge the gap between languages while avoiding overfitting even with limited parallel data. We consider simple (weighted) averaged word embeddings for a source language sentence as input but generates a sentence embedding in a target language space resulting from (weighted) averaged word vectors in a target language. The bi-directional structure additionally enables joint transformations between two languages.

The major contributions of the paper can be outlined as follows:
1) We highlight the simplicity and effectiveness of inducing high-quality sentence representations from pre-trained word embeddings by means of Term Frequency-Inverse Document Frequency (TF-IDF) weighted averaging word vectors.
2) We propose an adversarially bidirectional model for cross-lingual sentence embedding mapping based on a custom form of GAN framework that is capable of utilizing non-parallel sentence pairs. 
Moreover, we show that the same model architecture can easily be extended to more than one source language.
3) We extensively evaluate the performance of our method on the Tatoeba and Europarl corpora, obtaining exceptional accuracy as well as high quality mapping results, even in low-resource settings.

\section{Related Work}
\label{sec:related-work}

\paragraph{Cross-Lingual Projection Approaches.}
A number of papers consider linear projections to align two word vector spaces with a regression objective \cite{Mikolov2013ExploitingSA,zou2013bilingual}. 
\newcite{FaruquiDyer:2014:EACL} proposed using Canonical Correlation Analysis (CCA). 
\newcite{Xing2015NormalizedWE} showed that adding an orthogonality constraint to the mapping can significantly enhance the result quality, and has a closed-form solution.  
There have been approaches that assume that languages share some common vocabulary items as a heuristic for supervision  \cite{Smith2017OfflineBW,dong2018cross,artetxe2017learning}.

A few works also attempt to align monolingual word vector spaces with no supervision at all. 
\newcite{P17-1179} employed a form of adversarial training, but their approach differs from ours in multiple respects. First, they rely on sharp drops of discriminator accuracy for model selection. Second, their performance is highly sensitive to the selected parallel corpus. \newcite{lample2018word} presented a related unsupervised technique that learns a rotation matrix that outperforms several state-of-the-art supervised techniques. 
In contrast to our approach, none of the above methods consider non-linear transformations \cite{Nakashole2018}.

\paragraph{Sentence Embeddings.} Well-known approaches to create sentence embeddings include the Paragraph Vector approach \cite{LeMikolov2014ParagraphVectorPMLR}, which straightforwardly extends word2vec to generate vectors for paragraphs, and the Skip-Thought Vector approach \cite{Kiros2015SkipThoughtVectors}, which relies on recurrent units to encode and decode sentence representations such that these are predictive of neighbouring sentences. There are more sophisticated methods that rely on supervision from a range of different NLP tasks
\cite{SubramanianEtAl2018MultiTaskSentenceEmbeddings,Yang2019ImprovingMS}. 

However, 
inspired by the results from \newcite{wieting2015towards},
\newcite{arora2017asimple} presented a weighting technique that enables simple weighted sums of word vectors to outperform several state-of-the-art models. In our experiments, we build on these insights and as well consider weighted sums of word vectors as sentence embeddings, as these are readily available, even for many low-resource languages. Our weighting scheme is described in Section 
\ref{sec:eval}.

\vspace{-10pt}
\paragraph{Adversarial Training.}
Recently, GANs \cite{Goodfellow2014GANs} have shown remarkable success across a diverse range of multimodal tasks.
Their adversarial training process
resembles a min--max game.
Some GAN approaches require a supervised learning setting like image-to-image transfer \cite{pix2pix2017}. The CycleGAN approach \cite{zhu2017unpaired} shows promise in its exploitation of unpaired data to achieve a domain transfer. While GANs have mostly been considered for multimodal data \cite{liu2019oogan}, we show how they can be used for linguistic representations in an NLP task.

\section{\ourmethod\ Approach}

In this paper, we seek to learn a transformation between two languages such that the mapping model can be invoked to project an embedding of a source language sentence to a target language space and be able to find the nearest neighbor targets in the target space at the sentence level. At the same time, our approach is shown to be robust under low-resource conditions in terms of the amount of parallel sentences available for training.
We start with a formal definition of our sentence representation problem under limited parallel data, followed by a detailed demonstration of our proposed deep neural model.

\subsection{Problem Definition}
Formally, we assume a source language (domain) $X$ and a target one $Y$, such that each element is a $d$-dimensional vector, denoted by $\mathbf{x}\in X$ or $\mathbf{y}\in Y$.
We assume that 
$\forall \mathbf{x}\in X$, $\mathbf{x}$ is aligned with one 
$\mathbf{y}\in Y$, denoted by $(\mathbf{x},\mathbf{y})\in\mathcal{D}$.
Given a bilingual corpus $\mathcal{D}$ and a labeled (parallel) subset $\mathcal{D}_\mathrm{l}\subseteq\mathcal{D}$, an unlabeled subset $\mathcal{D}_\mathrm{ul}$ and some distance measure $f_{d}$, our goal is to learn two non-linear transformation functions $G_X$ and $G_Y$ that minimize
\begin{equation}
\label{eq:loss_dist}
\mathcal{L}_d = \mathbb{E}_{\mathbf{x},\mathbf{y}} \left[ f_{d}(\mathbf{x},G_Y(\mathbf{y}))+f_{d}(\mathbf{y},G_X(\mathbf{x})) \right]
\end{equation}
Note that in the labeled set $\mathcal{D}_\mathrm{l}$, the alignment between $\mathbf{x}$ is known, while in the unlabeled set $\mathcal{D}_\mathrm{ul}$, the relationship between $\mathbf{x}$ and $\mathbf{y}$ are unknown.

\begin{figure*}[htb]
\centering
\includegraphics[width=\textwidth]{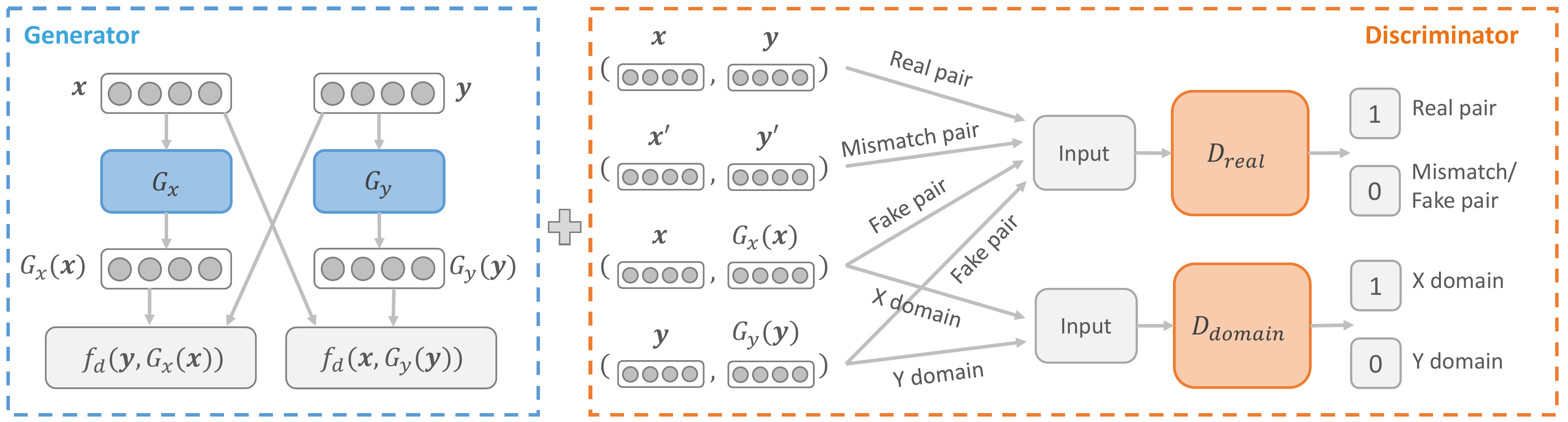}
\caption{The framework of our proposed \ourmethod\ method. It learns two generators $G_X$ and $G_Y$ to approximate the joint distribution of vectors from both languages. $G_X$ projects sentence embeddings $\mathbf{x}$ from language $X$ to $Y$, while, conversely, $G_Y$ projects sentence embeddings $\mathbf{y}$ from language $Y$ to $X$.}
\label{fig:model}
\end{figure*}

In this paper, we consider the case when the labeled set $|\mathcal{D}_\mathrm{l}|$ is very limited and the size of the unlabeled set $|\mathcal{D}_\mathrm{ul}|$ is large.
In other words, the challenge is how to utilize unpaired vectors from two domains to learn good mappings from one domain to the other. 
Specifically, the model is expected to be able to jointly learn from bidirectional transformations between the two languages at the same time to improve the mappings for each direction by better modeling the joint distribution, which is important in settings with limited parallel data as considered in this paper.

\subsection{Our Method}
\label{sec:method}
To solve the trasnformation problem with limited parallel data, we introduce the novel \emph{Adversarial Bi-directional Sentence Embedding Mapping} (\ourmethod) method.

The fundamental core of our model is inspired by the Triangle Generative Adversarial Network \cite{GanCWPZLLC17}, which addresses the task of image-to-image translation.
The key idea is that the generator component learns to non-linearly project embeddings across the two representation spaces, while a discriminator component attempts to distinguish automatically projected embeddings from genuine target language embeddings, thus constraining us to more closely match the target distribution.
Unlike regular GANs, our model incorporates additional information from ``adversarial'' pairs of sentence embeddings that come from both parallel and non-parallel data. We define the corresponding objective function as follows.
\begin{equation}
\label{eq:loss_real}
\begin{aligned}
  \mathcal{L}_\mathrm{real} & =~ 
  \mathbb{E}_{\mathbf{x},\mathbf{y}}[\log(D_\mathrm{real}(\mathbf{x}, \mathbf{y}))] ~ \\
  & + \mathbb{E}_{\mathbf{x}}[\log(1 - D_\mathrm{real}(\mathbf{x}, G_X(\mathbf{x}))] ~ \\
  & + \mathbb{E}_{\mathbf{y}}[\log(1 - D_\mathrm{real}(\mathbf{y}, G_Y(\mathbf{y}))]
\end{aligned}
\end{equation}
\noindent Here, $D_\mathrm{real}$ is a discriminator that aims to distinguish \emph{real pairs} from \emph{fake pairs}.
A \emph{real pair} $(\mathbf{x}, \mathbf{y})\in\mathcal{D}_l$ is a known mapping in the parallel dataset. 
A \emph{fake pair} $(\mathbf{x},G_X(\mathbf{X}))$ or $(\mathbf{y}, G_Y(\mathbf{Y}))$ is an artificial pair based on a projection emitted by the generator $G_X$ or $G_Y$.

Equation \ref{eq:loss_real} reflects an adversarial min--max game, in which the generators $G_X$, $G_Y$ and the discriminator $D_\mathrm{real}$ are trained adversarially and concurrently to improve their respective abilities. This is a  bidirectional process due to its reliance on both generator functions $G_X$ and $G_Y$ to map from one domain to the other.

In addition, in order to utilize non-parallel information, we further take into consideration \emph{mismatch pairs} induced from non-parallel data.
Given the set of source language sentence embeddings $X$ and the set of target language sentence embeddings $Y$, the set of \emph{mismatch pairs} consists of all training pairs $(\mathbf{x}',\mathbf{y}') \in X \times Y$
such that $\mathbf{x'}$ is an embedding for a sentence that is \emph{not} translationally equivalent to the sentence represented by $\mathbf{y'}$. The loss function is augmented with the following mismatch term:
\begin{equation}
\label{eq:loss_mis}
    \begin{aligned}
    	\mathcal{L}_{\mathrm{mis}} & =~  {\mathbb{E}_{\mathbf{x}',\mathbf{y}'}}[\log(1 - D_\mathrm{real}(\mathbf{x}',\mathbf{y}')] \\
    \end{aligned}
\end{equation}
By combining Equations \ref{eq:loss_real} and \ref{eq:loss_mis}, we force the discriminator $D_\mathrm{real}$ to distinguish \emph{real pairs} $(\mathbf{x},\mathbf{y})$ from unpaired data, which includes both \emph{mismatch pairs} $(\mathbf{x'}, \mathbf{y'})$ as well as generated \emph{fake pairs} $(\mathbf{x}, G_X(\mathbf{x}))$ or $(\mathbf{y}, G_Y(\mathbf{y}))$. 

However, the discriminator $D_\mathrm{real}$ alone cannot determine the directionality between \emph{fake pairs}.
Therefore, we introduce another discriminator $D_\mathrm{dom}$ to distinguish whether fake pairs come from the $X$ domain or from the $Y$ domain. The loss function is defined as: 
\begin{equation}
    \begin{aligned}
    	\mathcal{L}_{\mathrm{dom}} & =~  
    	\mathbb{E}_{\mathbf{x}}[\log(D_\mathrm{dom}(\mathbf{x},G_X(\mathbf{x}))] ~ \\
     	& +\mathbb{E}_{\mathbf{y}}[\log(1 - D_\mathrm{dom}(\mathbf{y}, G_Y(\mathbf{y}))]\\
    \end{aligned}
\end{equation}
The overall framework of our \ourmethod\ method is illustrated in Figure \ref{fig:model}, where we seek to solve the following joint optimization problem:
\begin{equation}
\label{eq:loss_joint}
\begin{aligned}
\mathcal{L} =~ & \mathcal{L}_\mathrm{real} + \mathcal{L}_{\mathrm{mis}} + \mathcal{L}_{\mathrm{dom}}\\
G_X^*, G_Y^* =~ &  \arg\mathop{\min}\limits_{G_{X}, G_Y}\mathop{\max}\limits_{D_\mathrm{real},D_\mathrm{dom}}(\mathcal{L} + \lambda \mathcal{L}_{d})
\end{aligned}
\end{equation}
where $\lambda$ is a weighting factor to balance the effect between distance metrics and adversarial components.
In this paper, we use cosine similarity as the distance measure:
\begin{equation}
\begin{aligned}
f_{\mathrm{d}}(\mathbf{x}, \mathbf{x'}) & =~ 1 - \frac{\mathbf{x}^\intercal \mathbf{x'}}{\left \| \mathbf{x} \right \| \, \| \mathbf{x'} \| } ~\\
\end{aligned}
\end{equation}
We train our model adversarially to learn the mappings bi-directionally by encouraging that the resulting pairs be indistinguishable from genuine pairs, and the direction that was generated remain as indiscernible as possible.

\subsection{Zero-Shot Multilingual Setting}
Our model can also be easily extended to align sentences between two languages $X_1$ and $X_2$ in a zero-shot manner without any parallel data between them.
The zero-shot multilingual task involves jointly projecting two languages $X_1$ and $X_2$ to a common target language $Y$ given only limited parallel data and non-aligned data connecting each to $Y$. As input, we have labeled data $\mathcal{D}_\mathrm{l}^{(i)}$ and unlabeled data $\mathcal{D}_\mathrm{ul}^{(i)}$ for the language pairs $(X_i,Y)$ ($i=1,2$).
However, we do not observe any direct relationship between the two source languages $X_1$ and $X_2$ in the training data.

In this case, we adopt the same framework as in the previous section except that the generator $G_x$ is expected to learn two language projections from $X_1$ to $Y$ and from $X_2$ to $Y$.
Let $\mathcal{L}^{(i)}$ be the loss function defined in Equation \ref{eq:loss_joint} for cross-lingual $X_i$ and $Y$.
The overall loss function for multilingual mapping is simply 
$\mathcal{L}' = \frac{1}{2}\left(\mathcal{L}^{(1)}+\mathcal{L}^{(2)}\right)$.

\subsection{Sentence Representation and Mapping}
\label{sec:sent-rep}

With regard to obtaining the sentence representations, we adopt pre-trained word vectors \cite{bojanowski2016enriching} that are available for numerous languages trained on Wikipedia using fastText. We ensure the same input embeddings in training and evaluation for the baselines as in our model.
Based on the results of \newcite{wieting2015towards} and \newcite{arora2017asimple}, we adopt simple (weighted) averages of word vectors, which are surprisingly powerful, although our method could also be applied to other sentence embedding methods.

Given source sentence embeddings $\{\mathbf{x}\}$ and target sentence embeddings $\{\mathbf{y}\}$ acquired as described above, we can train the generators $G_X$ and $G_Y$ through the joint loss function in Equation \ref{eq:loss_joint}.
Subsequently, we evaluate the obtained transformation via a standard sentence retrieval task. 
For each source sentence embedding $\mathbf{x}$, 
we compute its $k$ nearest neighbours in terms of the distance function $f_d$ among all target embeddings.
The corresponding $k$ target sentences are regarded as the candidate set of mapping results.

\newcommand{\zerosp}{}
\begin{table*}[t]
\centering
\small
\begin{tabular*}{\textwidth}{l|p{1.58em}p{1.58em}p{1.58em}p{1.58em}p{1.58em}p{1.58em}p{1.58em}p{1.58em}|p{1.58em}p{1.58em}p{1.58em}p{1.58em}p{1.58em}p{1.58em}p{1.58em}p{1.58em}}
\hline
Datasets & \multicolumn{8}{c|}{\textbf{Tatoeba}} & \multicolumn{8}{c}{\textbf{Europarl}} \\ 
Corpus & \multicolumn{2}{c}{deu$\rightarrow$eng} & \multicolumn{2}{c}{eng$\rightarrow$deu} & \multicolumn{2}{c}{spa$\rightarrow$eng} & \multicolumn{2}{c|}{eng$\rightarrow$spa} & \multicolumn{2}{c}{deu$\rightarrow$eng} & \multicolumn{2}{c}{eng$\rightarrow$deu} & \multicolumn{2}{c}{spa$\rightarrow$eng} & \multicolumn{2}{c}{eng$\rightarrow$spa} \\
\hline
Precision@$k$ (\%) & $k$=1 & $k$=5 & $k$=1 & $k$=5 & $k$=1 & $k$=5 & $k$=1 & $k$=5 & $k$=1 & $k$=5 & $k$=1 & $k$=5 & $k$=1 & $k$=5 & $k$=1 & $k$=5 \\
\hline
Mikolov et al.~\shortcite{Mikolov2013ExploitingSA}
& 13.6 & 20.9 & 12.8 & 21.7 & 31.1 & 46.9 & 24.7 & 38.3 & \zerosp 6.4 & 13.4 & \zerosp 7.8 & 14.6 & 13.1 & 21.6 & 12.4 & 22.1 \\
Dinu et al.~\shortcite{dinu2014improving}
& 17.3 & 30.8 & 22.7 & 36.4 & 35.4 & 52.8 & 30.5 & 46.2 & 13.8 & 21.9 & 11.9 & 21.1 & 18.5 & 28.6 & 15.7 & 27.3 \\
\protect\newcite{Smith2017OfflineBW} & 27.6 & 42.6 & 26.0 & 40.0 & \textbf{40.5} & 55.2 & 35.0 & 48.1 & 13.9 & 24.3 & 19.3 & 27.5 & 47.2 & 58.6 & 39.5 & 50.8 \\
\protect\newcite{lample2018word} & 10.0 & 19.7 & 11.2 & 18.7 & 29.1 & 40.3 & 30.5 & 43.5 & 20.8 & 30.2 & 22.9 & 35.2 & 28.2 & 39.7 & 30.0 & 42.7 \\
BERT                           & 10.0 & 17.3 & 12.4 & 21.1 & 25.6 & 38.2 & 24.9 & 33.8 & \zerosp 5.9  & \zerosp 8.9  & \zerosp 5.8  & \zerosp 9.6 & 22.5 & 39.9 & 21.7 & 36.5 \\
seq2seq NMT                    &  \zerosp 6.5 &  \zerosp 9.0 &  \zerosp 7.3 & 10.1 &  \zerosp 9.9 & 12.5 &  \zerosp 9.9 & 13.1 & \zerosp 5.0  & \zerosp 8.7  & \zerosp 5.1  & \zerosp 6.3 & \zerosp 7.0  & \zerosp 9.2  & \zerosp 7.8 & 10.3\\
fairseq NMT                    &  \zerosp 9.3 & 19.4 & 15.6 & 24.4 & 19.9 & 29.8 & 18.5 & 31.0 & 12.2 & 16.1 & \zerosp 7.3 & 11.8 & 18.4 & 27.9 & 17.0 & 25.2 \\
Conditional GAN                & 16.0 & 36.3 & 18.8 & 37.5 & 16.6 & 34.2 & 16.6 & 32.3 & \zerosp 9.6 & 17.1 & \zerosp 7.9 & 15.0 & 18.6 & 29.6 & 17.3 & 27.6 \\
\ourmethod\ (ours) & \textbf{46.2} & \textbf{65.5} & \textbf{43.8} & \textbf{65.9} & 38.5 & \textbf{64.3} & \textbf{37.4} & \textbf{60.5} & \textbf{27.5} & \textbf{40.4} & \textbf{26.2} & \textbf{39.2} & \textbf{49.4} & \textbf{62.5} & \textbf{47.7} & \textbf{62.2} \\
\hline
\end{tabular*}
\caption{Results of sentence embedding mapping experiment in terms of precision@1 and precision@5. Our proposed \ourmethod\ only utilizes 20\% of parallel training data and the equal size of unparalleled data for training, while all baselines take 100\% training data with parallel labels. The best-performing method is highlighted in bold. 
}
\label{tab:main_exp}
\end{table*}

\section{Experiments}
\label{sec:eval}
In this section, we extensively evaluate the effectiveness of our \ourmethod\ method compared with state-of-the-art approaches on two heterogeneous real-world corpora.

\subsection{Experimental Setup}
\label{sec:experimental_setup}

\paragraph{Datasets.}

We evaluate the precision of our approach 
on the Europarl parallel corpus and on extracted from 
the Tatoeba service\footnote{http://tatoeba.org}, which provides translations of commonly used phrases that might be useful to language learners. We focus on German and English as well as Spanish and English translation retrieval. 
For the English $\longleftrightarrow$ German datasets, we take 160k pairs as the training set and 1,600 pairs as the test set in both datasets.
For the English $\longleftrightarrow$ Spanish datasets, we take 60k pairs as training and 600 pairs as test data for the Tatoeba corpus, and 130k as training and 1,300 as test for the Europarl corpus.
However, to emphasize that our model can cope with very limited amounts of parallel data, we solely make use of just 20\% of the parallel training data when training our model, while all the baseline methods exploit 100\% of the parallel training data. Given a set of training pairs, we randomly sample false pairs of the same size as the respective parallel data.

\vspace{-10pt}
\paragraph{Baselines.}

For comparison, we consider as baselines the linear transformation methods by \cite{Mikolov2013ExploitingSA}, \newcite{dinu2014improving}, and \newcite{Smith2017OfflineBW}, the supervised version of MUSE \cite{lample2018word} with cross-domain similarity local scaling. We also use the multilingual version of BERT \cite{devlin2018bert}, 
using the standard method for sentence-level representations based on the [CLS] token\footnote{As provided by based on 
bert-as-service: \url{https://github.com/hanxiao/bert-as-service}} to generate sentence vectors that are already multilingual without further projection. This is to assess how far we can take simple word vector averages in comparison to powerful alternatives.

We further consider a seq2seq \cite{sutskever2014sequence} NMT baseline jointly trained to translate language $X$ to $Y$ as well as to monolingually auto-encode sentences from language $Y$ back to $X$. We use two different encoders with a shared decoder such that the two encoders produce latent representations in the same space. This allows us to save the latent sentence embeddings for evaluation rather than generate an output translation.

Additionally, we consider the fairseq NMT \cite{gehring2017convs2s} 
approach based on a convolutional encoder model, which constructs latent representations hierarchically.

Finally, we investigate a Conditional GAN \cite{Mirza2014ConditionalGA}, for which we use our model but do not consider any fake pairs or mismatch pairs for training.

\vspace{-10pt}
\paragraph{Parameter Settings.}
Both generators $G_X$ and $G_Y$ consist of three fully connected layers with hidden sizes of 512, 1024, 512, respectively. Each hidden layer is connected with a \emph{BatchNorm} layer and the \emph{ReLU} activation function. The final activation function is \emph{tanh}. Both discriminators $D_\mathrm{real}$ and $D_\mathrm{dom}$ take as input two embeddings, followed by three fully connected layers of sizes 512, 1024, 512 with concatenation. Each hidden layer is connected with a \emph{leaky ReLU} activation function (0.2), while the output is activated by a \emph{sigmoid} function.
We rely on Adam optimization with an initial learning rate of 0.002 and a batch size of 128.

\subsection{Main Results}
We assess the quality of the mapping by considering the ranking of the ground truth paired target sentence. The overall quality across all test set instances is given by the precision@$k$ metric, which, following previous work \cite{Mikolov2013ExploitingSA} in this area, is defined as the ratio of test set instances for which the correct target is among the top $k$. Then we repeat the same evaluation process for all four datasets from the two corpora. 

The results are reported in Table \ref{tab:main_exp}. Recall that we only use 20\% of parallel sentences (true pairs) to train our model while all the other baselines utilize 100\% of parallel sentence pairs for the training. 
We observe that our \ourmethod\ approach still significantly outperforms other baselines by a large margin.
Take the deu$\rightarrow$eng data from Tatoeba as an example. Our method achieves a precision@1 of 46.2\% and a precision@5 of 65.5\%, which are 18.6 and 22.9 \emph{absolute} percentage points higher than the respective results of the best baseline.
Similar trends can be observed for other datasets. Note that the results for different languages are not fully comparable due to different sizes of  training data. 

\begin{figure*}[t]
\centering
\includegraphics[width=0.24\textwidth]{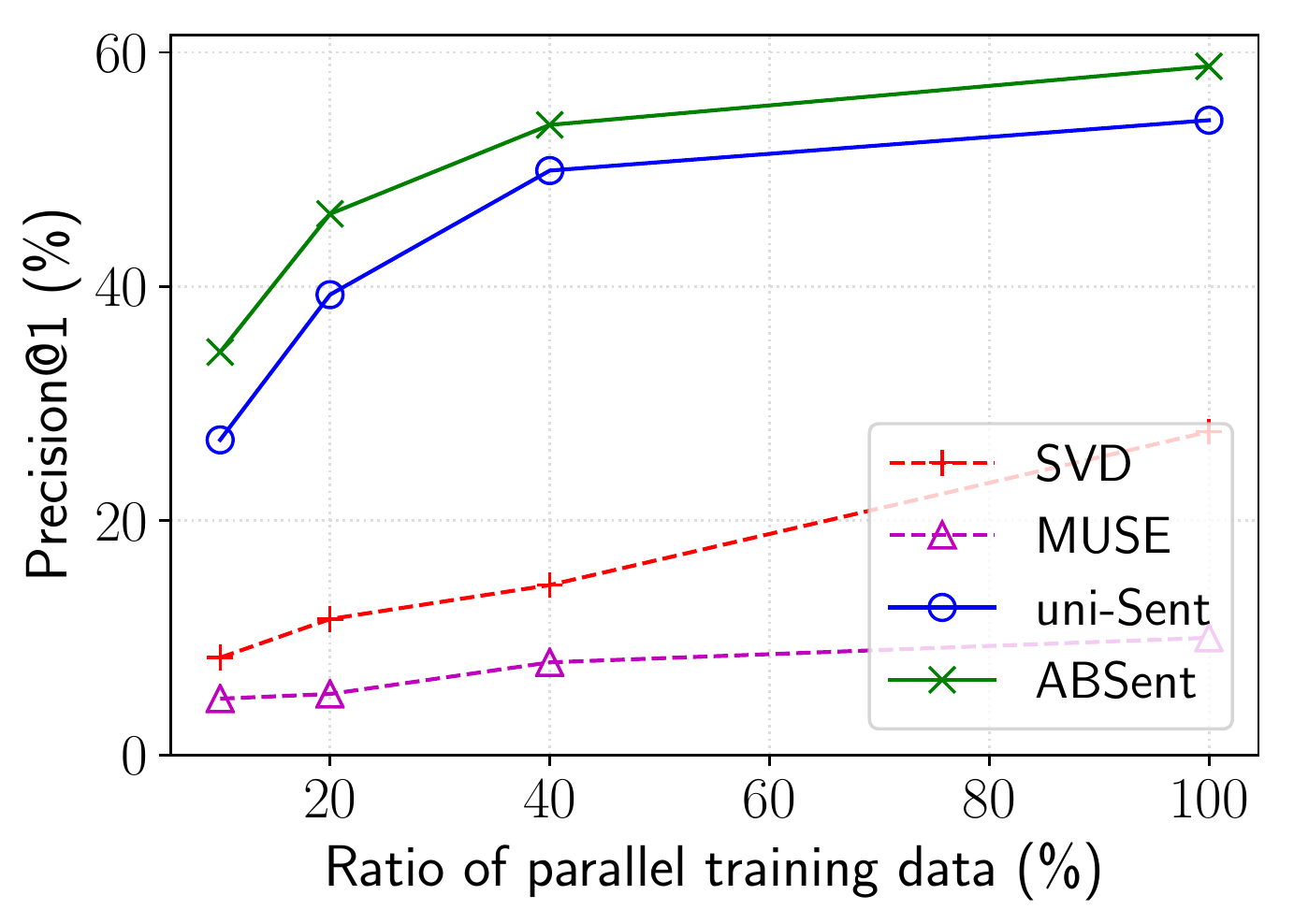}
\includegraphics[width=0.24\textwidth]{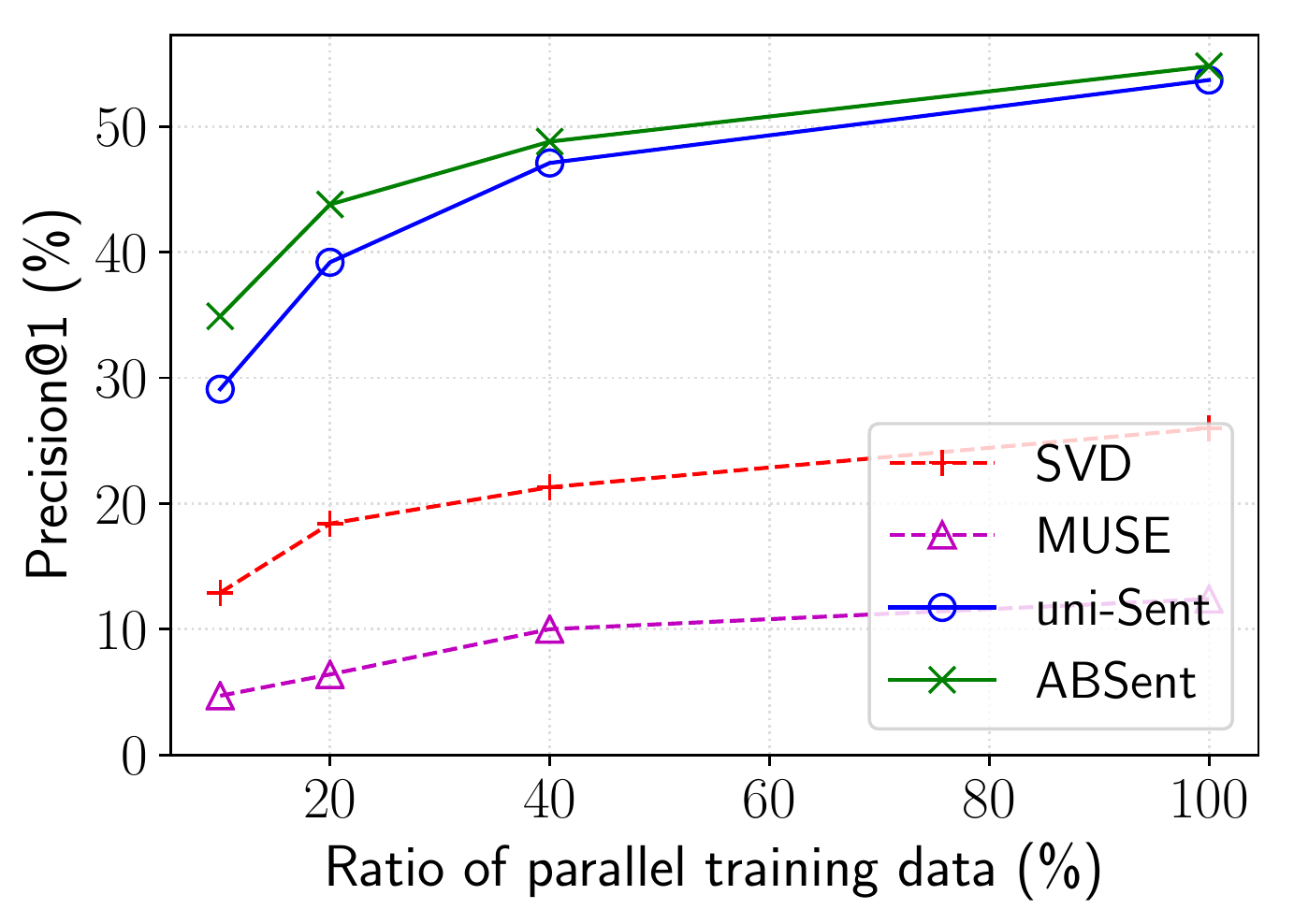}
\includegraphics[width=0.24\textwidth]{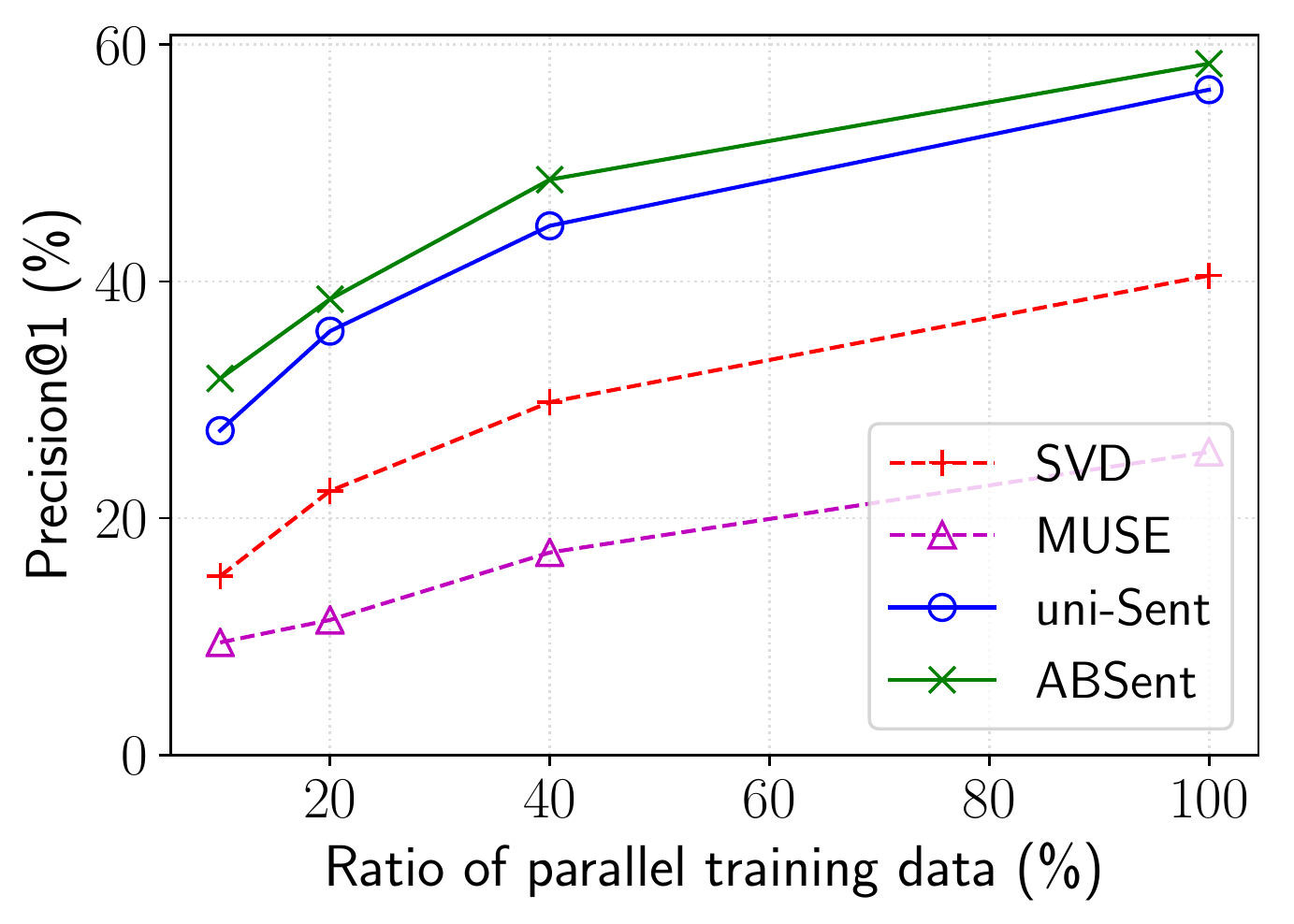}
\includegraphics[width=0.24\textwidth]{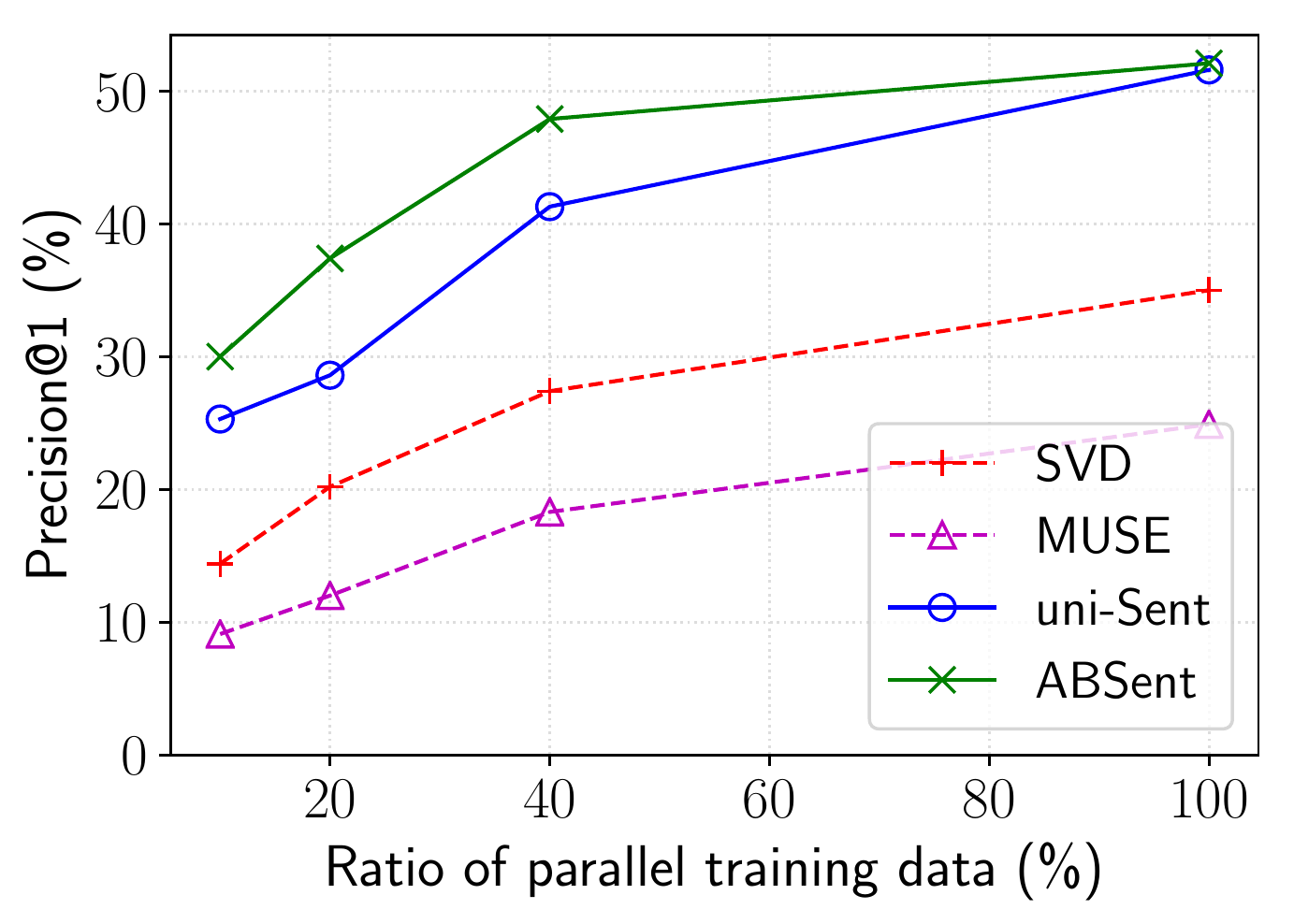} \\
\small \hspace{0.2in} (a) deu$\rightarrow$eng \hspace{1.08in} (b) eng$\rightarrow$deu \hspace{1.08in} (c) spa$\rightarrow$eng \hspace{1.08in} (d) eng$\rightarrow$spa \\
\includegraphics[width=0.24\textwidth]{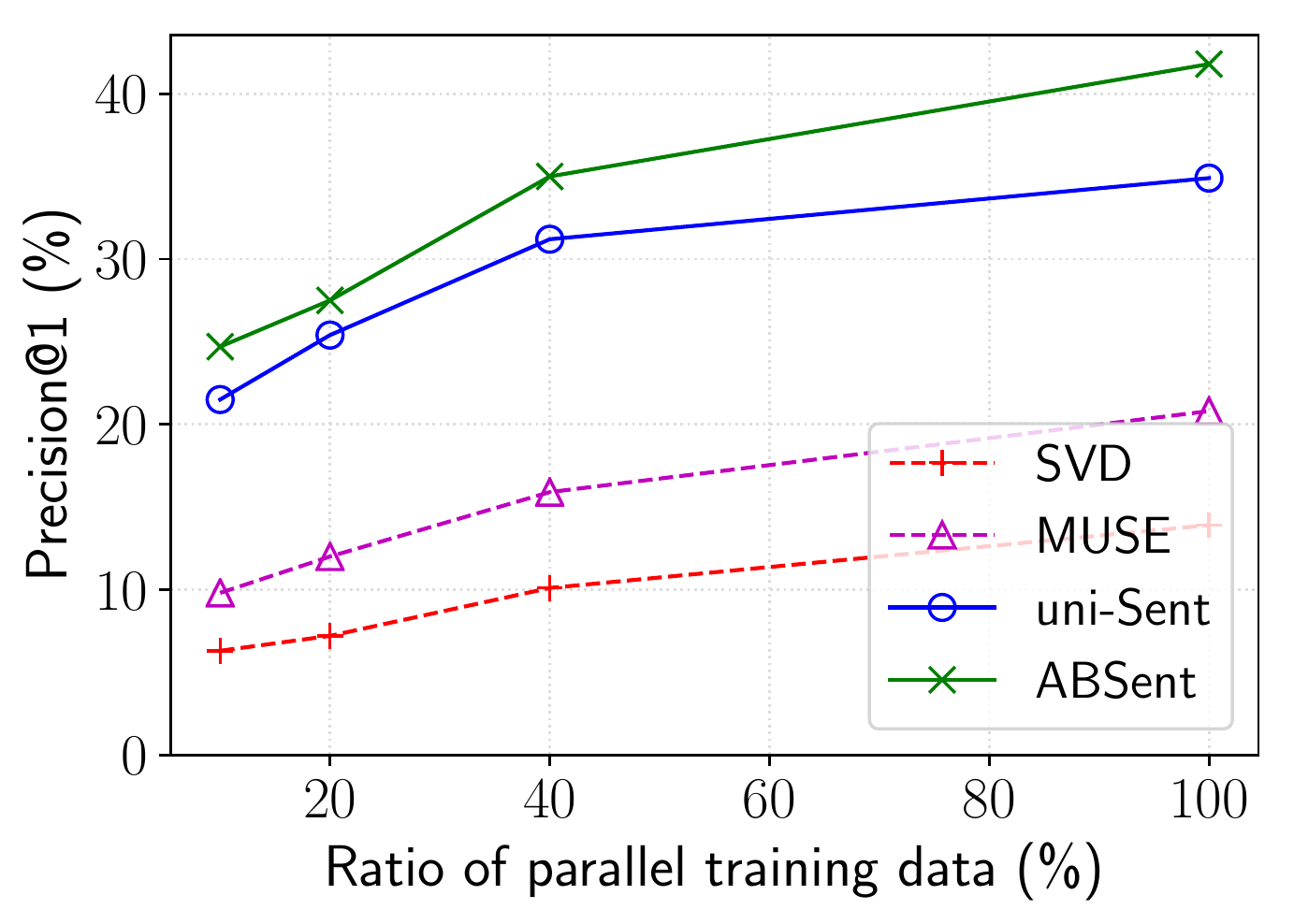}
\includegraphics[width=0.24\textwidth]{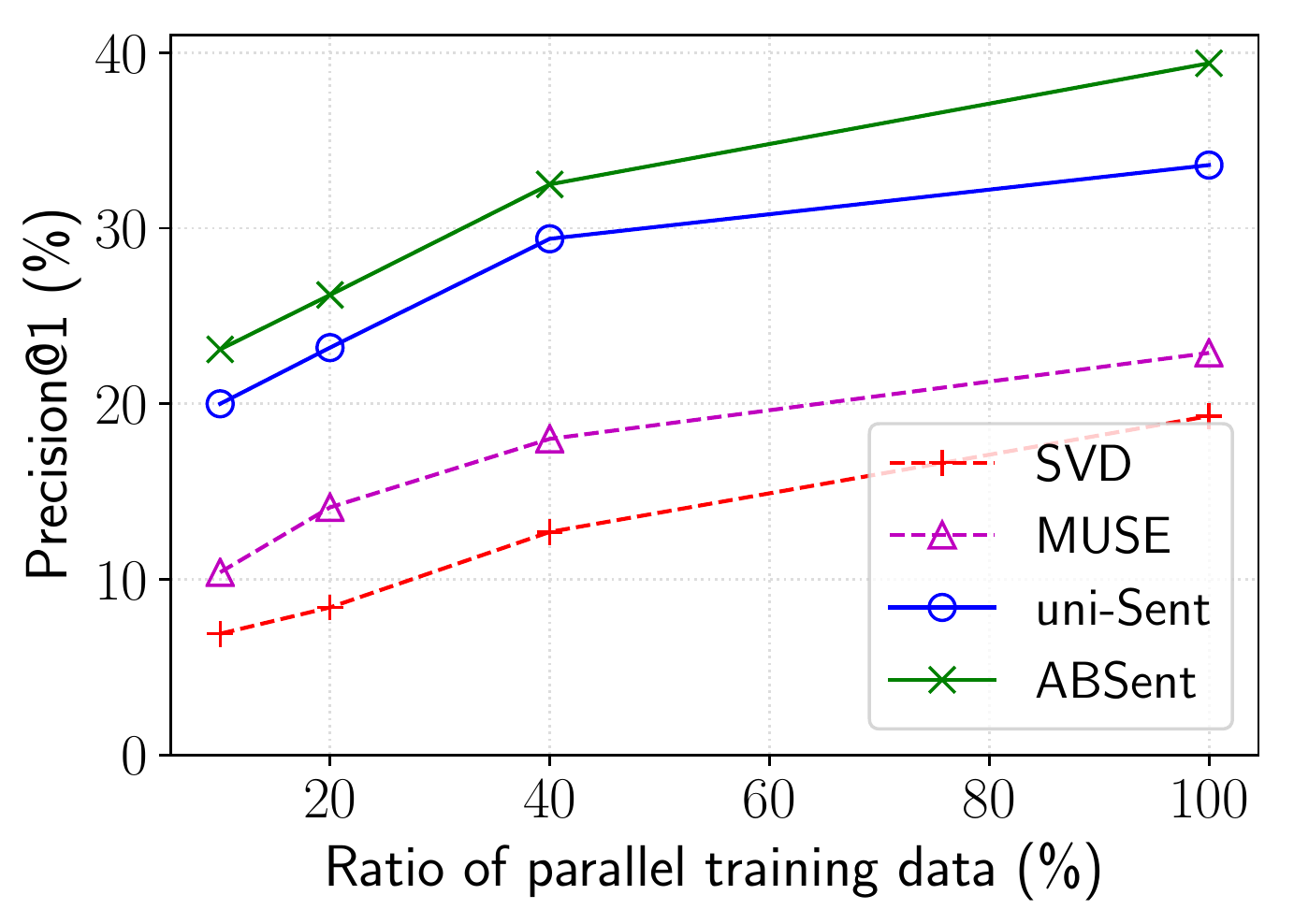}
\includegraphics[width=0.24\textwidth]{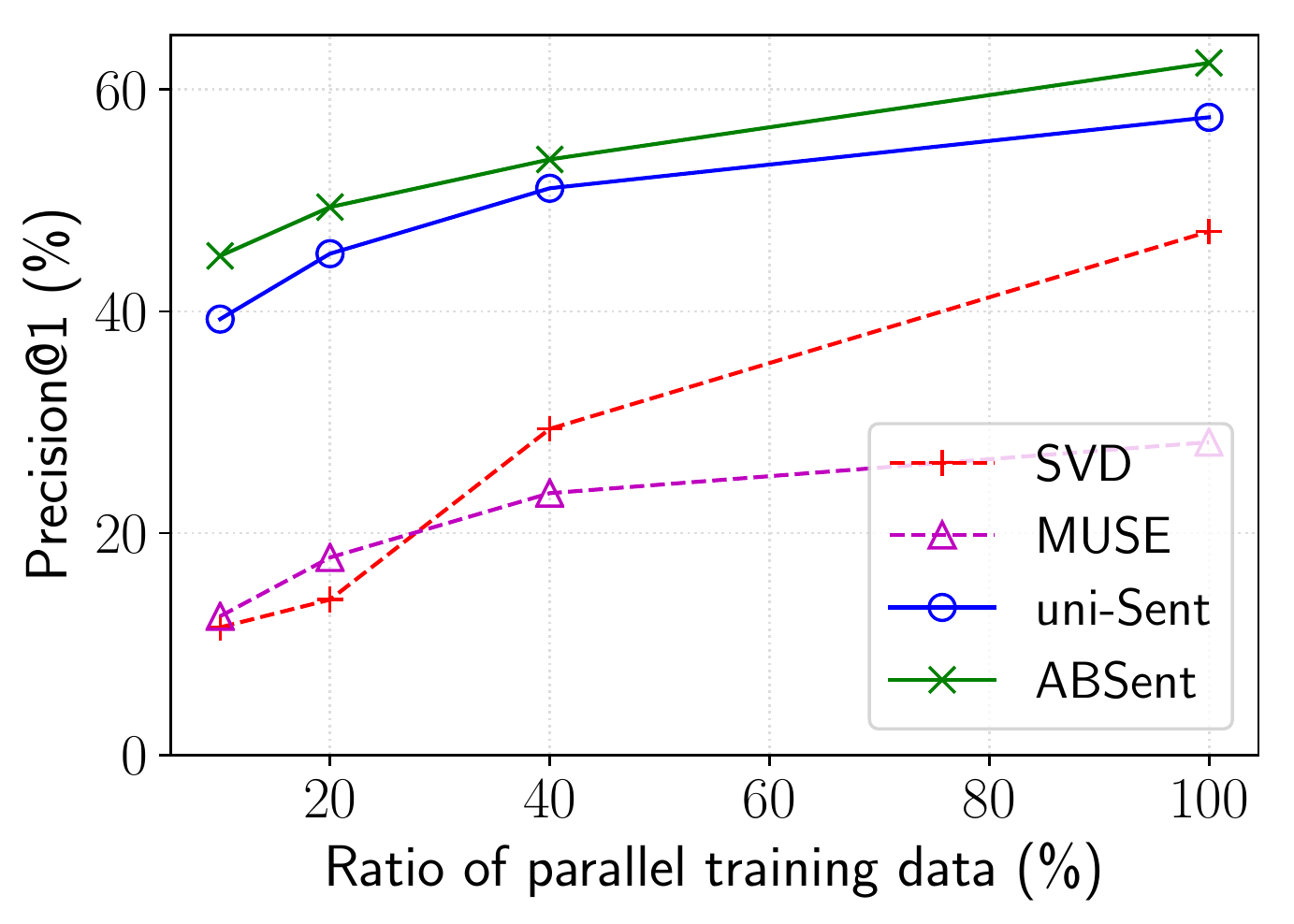}
\includegraphics[width=0.24\textwidth]{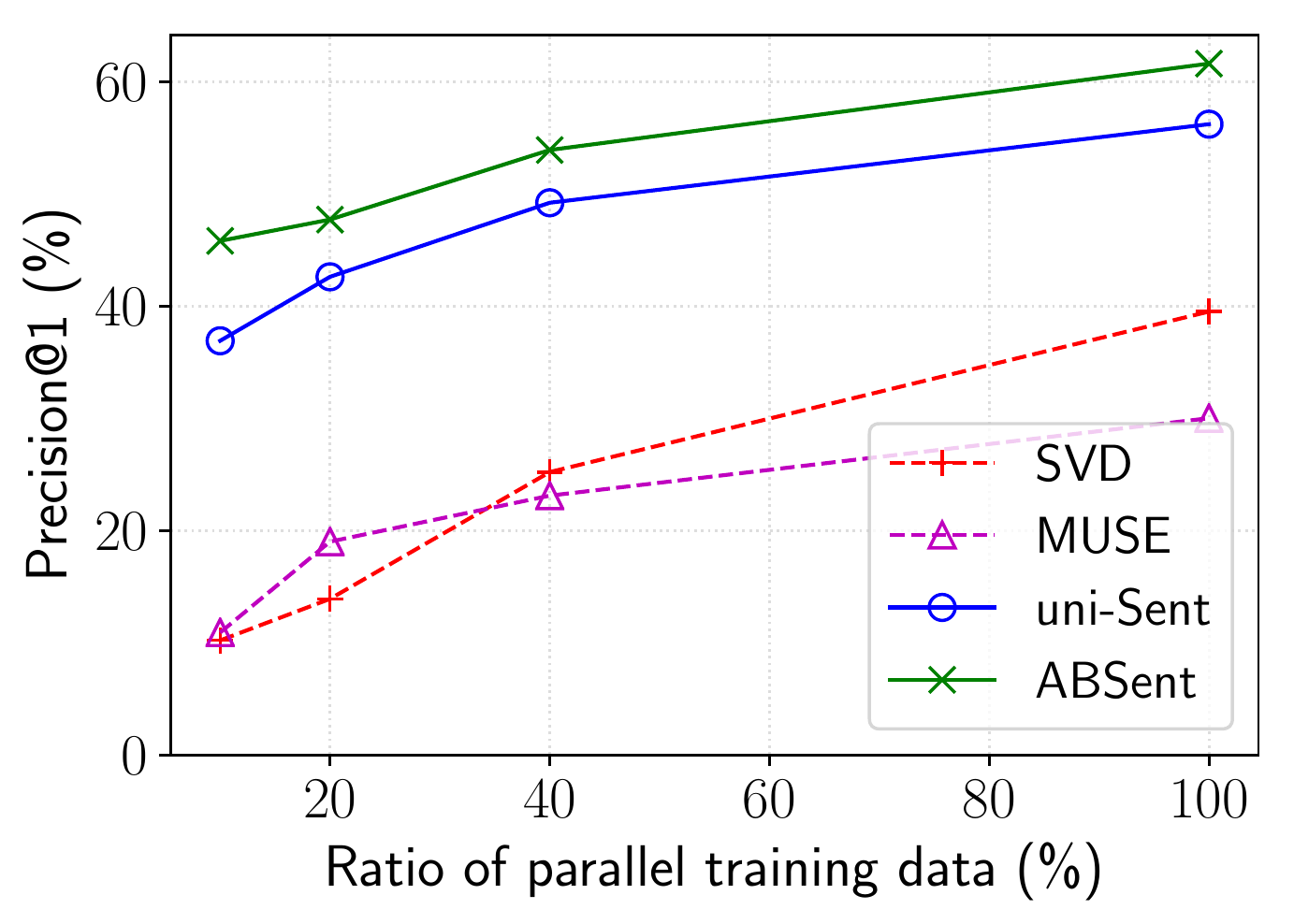} \\
\small \hspace{0.2in} (e) deu$\rightarrow$eng \hspace{1.08in} (f) eng$\rightarrow$deu \hspace{1.08in} (g) spa$\rightarrow$eng \hspace{1.08in} (h) eng$\rightarrow$spa \\
\caption{Performance of our proposed \ourmethod\ and \ourmethoduni\ with two baseline experiments under different ratios of parallel training data on the Tatoeba dataset (Figures (a)-(d)) and Europarl dataset (Figures (e)-(h)).}
\label{fig:exp_bi_tatoeba}
\end{figure*}

\subsection{Detailed Analysis}

\paragraph{Influence of Bi-directional Transformation.}
We evaluate how the bi-directional mapping strategy affects the effectiveness of our model. Taking German sentence embeddings as source and English sentence embedding as the target, we train our model to map the German sentence embeddings to the corresponding English vector space and align the sentence with the same meaning. We obtain the unidirectional transformation model from German to English, which we refer to as \emph{\ourmethoduni}.
Then we repeat the same process for English to German, Spanish to English, and English to Spanish.
Thus, this model only acquires the ability to conduct a unidirectional transformation between two languages, since the bi-directional discriminator $D_\mathrm{dom}$ is omitted. In this case, we only learn the generator $G_X$ to map the source to the target domain. 
The goal is to optimize
\begin{equation}
    \begin{aligned}
     	\mathcal{L}_{\mathrm{real}} & ~=~ \mathbb{E}_{\mathbf{x},\mathbf{y}}[\log(D_\mathrm{real}(\mathbf{x}, \mathbf{y}))] \\
     	& ~+~ \mathbb{E}_{\mathbf{x}}[\log(1 - D_\mathrm{real}(\mathbf{x},{G(\mathbf{x})}))] \\
    	 G_X^* & ~=~ \mathop{\min}\limits_{G_X}\mathop{\max}\limits_{D_\mathrm{real}}(\mathcal{L}_\mathrm{real}+\mathcal{L}_\mathrm{mis} + \lambda \mathcal{L}_{d}) \\
    \end{aligned}
\end{equation}
where  $\mathcal{L}_{\mathrm{mis}}$ and $\mathcal{L}_d$ stay the same as in Equations \ref{eq:loss_mis} and \ref{eq:loss_dist}.
The results in Figure \ref{fig:exp_bi_tatoeba} shows that
the effectiveness of the method is improved substantially by the introduction of bidirectional learning. This demonstrates that the bidirectional training method not only enables a simultaneous transformation between the two language sentence embeddings, but also delivers better results. One possible reason is the effectiveness of $D_\mathrm{dom}$, which can regularize the directional uncertainty among two language sentence embedding transformations. A unidirectional transformation does not bear this benefit.

\begin{table}[t]
\centering
\small
\setlength\tabcolsep{3pt}
\begin{tabular*}{\linewidth}{p{3.1em}|p{2.2em}p{2.2em}p{2.2em}p{2.2em}|p{2.2em}p{2.2em}p{2.2em}p{2.2em}}
\hline
Dataset & \multicolumn{4}{c|}{\textbf{Tatoeba}} & \multicolumn{4}{c}{\textbf{Europarl}} \\ 
Corpus & deu$\rightarrow$ & eng$\rightarrow$ &  spa$\rightarrow$ & eng$\rightarrow$ & deu$\rightarrow$ &  eng$\rightarrow$ &  spa$\rightarrow$ &  eng$\rightarrow$ \\
& \multicolumn{1}{c}{eng} & \multicolumn{1}{c}{deu} & \multicolumn{1}{c}{eng} & \multicolumn{1}{c|}{spa} & \multicolumn{1}{c}{eng} & \multicolumn{1}{c}{deu} & \multicolumn{1}{c}{eng} & \multicolumn{1}{c}{spa} \\
\hline
$U\setminus$mis & \multicolumn{1}{c}{25.1} & \multicolumn{1}{c}{28.0} & \multicolumn{1}{c}{27.1} & \multicolumn{1}{c|}{24.4} & \multicolumn{1}{c}{19.8} & \multicolumn{1}{c}{19.6} & \multicolumn{1}{c}{38.6} & \multicolumn{1}{c}{36.7}  \\
$U$ & \multicolumn{1}{c}{26.9} & \multicolumn{1}{c}{29.1} & \multicolumn{1}{c}{27.4} & \multicolumn{1}{c|}{25.3} & \multicolumn{1}{c}{21.5} & \multicolumn{1}{c}{20.0} & \multicolumn{1}{c}{39.3} & \multicolumn{1}{c}{36.9}\\
$A\setminus$mis & \multicolumn{1}{c}{32.0} & \multicolumn{1}{c}{32.9} & \multicolumn{1}{c}{30.7} & \multicolumn{1}{c|}{29.6} & \multicolumn{1}{c}{23.9} & \multicolumn{1}{c}{22.8} & \multicolumn{1}{c}{43.0} & \multicolumn{1}{c}{44.5}\\
$A$ & \multicolumn{1}{c}{\textbf{34.4}} & \multicolumn{1}{c}{\textbf{34.9}} & \multicolumn{1}{c}{\textbf{31.8}} & \multicolumn{1}{c|}{\textbf{30.0}} & \multicolumn{1}{c}{\textbf{24.7}} & \multicolumn{1}{c}{\textbf{23.1}} & \multicolumn{1}{c}{\textbf{45.0}} & \multicolumn{1}{c}{\textbf{45.8}}\\
\hline
$A_s$ & \multicolumn{1}{c}{27.4} & \multicolumn{1}{c}{28.1} & \multicolumn{1}{c}{28.0} & \multicolumn{1}{c|}{26.2} & \multicolumn{1}{c}{15.6} & \multicolumn{1}{c}{14.5} & \multicolumn{1}{c}{24.6} & \multicolumn{1}{c}{22.8}  \\
\hline
$A_w$ & \multicolumn{1}{c}{24.0} & \multicolumn{1}{c}{23.7} & \multicolumn{1}{c}{22.5} & \multicolumn{1}{c|}{21.5} & \multicolumn{1}{c}{21.9} & \multicolumn{1}{c}{22.4} & \multicolumn{1}{c}{37.1} & \multicolumn{1}{c}{38.4}  \\
\hline
\end{tabular*}
\caption{
Comparison of precision@1 between \ourmethod\ ($A$), \ourmethoduni\ ($U$), and further modified settings. $A\setminus$mis,
$U\setminus$mis are two variants without mismatch loss term. $A_s$ swaps the weighting strategy (TF-IDF weighting for Tatoeba but not for Europarl). $A_w$ stands for first aligning word-level representations and then generating sentence embeddings.
}

\label{tab:exp_mismatch}
\end{table}

\vspace{-10pt}
\paragraph{Influence of Ratio of Parallel Training Corpus.} 
Next, we study how various ratios of available parallel data affect the effectiveness of our model.
Apart from using 20\% of parallel data in the training corpus, we also evaluate using 10\%, 40\%, and 100\% as ratios of parallel training sentence pairs. We randomly sample the mismatch pairs to be of equal size as the respective parallel data.
The results are also depicted in Figure \ref{fig:exp_bi_tatoeba}. We observe that the precision improves as the ratio of parallel labelled sentence increases.

\vspace{-10pt}
\paragraph{Influence of Mismatch Loss.}
\label{sec:mismatch}
Recall that our \ourmethod\ method incorporates a custom mismatch loss in Equation \ref{eq:loss_mis}. This experiment aims to study how this loss function influences the effectiveness of our proposed models. For simplicity, we refer to a model \emph{X} without the mismatch loss as \emph{X$\setminus$mis}, where \emph{X} can be either \emph{\ourmethoduni} or \emph{\ourmethod}.
The results in Table \ref{tab:exp_mismatch} demonstrate the effectiveness of the mismatch strategy of introducing false pairs. We take 10\% of parallel training pairs, leaving other parameters as in the main experiments. Generally, bringing in mismatch pairs into the training improves the effectiveness of both our \ourmethod\ model and the \ourmethoduni\ version.

\vspace{-10pt}
\paragraph{Influence of Weighting strategy}
For the input sentence embeddings of \ourmethod, we take the average of word embeddings and get normalized to obtain sentence representations for the Tatoeba corpus. For the Europarl corpus, we define the sentence embeddings to be the normalized sum over the word vectors multiplied with TF-IDF scores for a weighted average. As a comparison experiment, we swap the weighting stragy in $A_s$ while keeping other parameters as for \ourmethod, i.e., we impose TF-IDF weighted vectors for Tatoeba, while using vanilla word averages for Europarl.

The effectiveness of $A_s$ is shown in Table \ref{tab:exp_mismatch}. Choosing an appropriate weighting strategy boosts the experimental results. Since TF-IDF weights words in accordance with their assumed importance, for the Europarl corpus, the $2\sigma$ volume of sentence lengths falls between 35 and 60, while for the Tatoeba corpus it is between 8 and 16. In such circumstances, words that appear more frequently in long sentences, especially function words such as 'a', 'the', etc.\ ought to have a lower weight, while infrequent ones ought to have a higher weight.
For short sentences, it may make sense to even consider words with higher frequency, so as to not neglect their semantic contribution to the sentence. Thus, simple averaging works better than TF-IDF weighted averaging for the Tatoeba corpus.

\vspace{-10pt}
\paragraph{First Aligning Words.}

As we can see in Table \ref{tab:exp_mismatch}, the accuracy drops quite notably compared to the regular \ourmethod\ approach if we first align words and then create the sentence vectors. The only difference between $A_w$ and \ourmethod\ is that we first align individual word vectors using our method and then average them (TF-IDF weighted averaging for the Europarl corpus) to generate sentence embeddings in the target vector space. For this comparison, we take 10\% of parallel word pairs for training. We conjecture that this approach is less able to account for variation in the meaning of a word across different sentences.

\vspace{-10pt}
\paragraph{Qualitative Analysis.}
We additionally provide representative examples of nearest neighbours, showcasing typical high-quality, medium quality, as well as low-quality transformation results for English--German in Table \ref{tab:examples}.
The lower quality results in some cases highlight the limits of average word vector based embeddings, as they disregard word order and may lose semantically salient signals. This problem can be overcome by applying our method using more semantically sophisticated methods to obtain sentence embeddings. Although many such methods require extensive training and in some cases also rich supervision, the advantage of our model is that we can rely on just limited parallel data to project a resource-poor language into the embedding space of a resource-rich language such as English for which such sentence embedding methods are readily available.

\begin{table}[t]
\centering
\small
\setlength\tabcolsep{3pt}
\begin{tabular}{l|lr|lr|l}
\hline
Language & \multicolumn{2}{l|}{LASER} & \multicolumn{2}{l|}{\ourmethod\ (ours)} & 
\\ 
\hline
         & \#train & Acc (\%) & \#train & Acc (\%) & \#test \\
\hline
Armenian (hye) & 6k & \textbf{32.21} & 5.3k & 28.56 & 742\\
Irish (gle) & 732 & 4.20 & 700 & \textbf{5.23} & 1k\\
Kazakh (kaz) & 4k & 17.39 & 3.9k & \textbf{18.14} & 575\\
\hline
\end{tabular}
\caption{On low-resource languages, simple word vector averages obtain comparable results to the richly supervised LASER model, despite lack of word order information.
}
\label{tab:lowresource}
\end{table}

\begin{table*}[t]
\centering
\small
\begin{tabular*}{\textwidth}{p{0.32in}p{0.9in}p{5.4in}}
\toprule
Example & German Sentence & English Sentences ranked by similarity score \\
\midrule
\multirow{3}{0.5in}{Tatoeba\\(high-quality)} & \multirow{3}{1.0in}{Ich finde keine Worte.} 
    & (1) \textbf{I am at a loss for words. (0.852)} \\
    &  & (2) I just don't know what to say. (0.847) \\
    &  & (3) Tom has turned twenty. (0.836) \\
\midrule
\multirow{3}{0.5in}{Europarl\\(high-quality)} & \multirow{3}{1.0in}{Notwendig ist die Interoperabilit\"at der Set-Top-Boxen.} 
    & (1) \textbf{Set top boxes must be compatible with each other. (0.895)} \\
    &  & (2) People are showing great solidarity at local and regional level and help is being mobilised at national level. (0.858) \\
    &  & (3) I support the proposed deadline for the Commission being 30 September in proposed amendment. (0.846) \\
\midrule
\multirow{4}{0.5in}{Europarl (medium-quality)} & \multirow{4}{1.0in}{Zun\"achst ist es zu begrüßen, dass der Universaldienst zwar einfache, aber keine breitbandigen Internetanschlüsse umfassen soll.} 
    & (1) In the past, the international community has done itself credit by prohibiting anti-personnel mines on these grounds. It should now, by the same token, ban weapons containing depleted uranium. (0.866) \\
    & & (2) This is because it provides clarity and therefore does not expose public services to the attacks which would otherwise have been levelled at them. (0.863) \\
    & & (3) That does not mean, however, that we do not still see much room for improvement, as other speakers have pointed out as well, and some of our wishes have not been fulfilled. (0.854) \\
    & & (4) \textbf{Firstly, the fact that the universal service is to include simple but not broadband Internet connections is to be welcomed. (0.852)} \\
\midrule
\multirow{5}{0.5in}{Europarl\\(low-quality)} & \multirow{5}{1.1in}{Sie teilt die Auffassung, dass mit dem Universaldienst nicht nur die geographische Abdeckung gewährleistet werden soll.} 
    & (1) At the European Council in Gothenburg at the end of this week, the Swedish Presidency will point out the need to discuss these issues within the European Union in order to develop a concrete basis allowing powerful action by the European Union on these vital issues. (0.862)\\
    & & (2) It could have expressed a lot more in the way of hopes for the future. (0.860)\\
    & & (3) So what does this railway package contain? (0.859) \\
    & & $\cdots$ \\
    & & (6) \textbf{The Commission shares the view that universal service is not just about getting geographical coverage right. (0.814)}\\
\bottomrule
\end{tabular*}
\caption{\label{tab:examples}Examples of English sentences as dom neighbors for German sentences in Europarl and Tatoeba. (parallel ratio=$10\%$)}
\end{table*}

\begin{table}[ht]
\centering
\small
\begin{tabular*}{\linewidth}{p{3.8em}|p{1.3em}p{1.3em}p{1.3em}p{1.5em}|p{1.3em}p{1.3em}p{1.3em}p{1.3em}}
\hline
\hspace{-5px}Datasets & \multicolumn{4}{c|}{\textbf{Tatoeba}} & \multicolumn{4}{c}{\textbf{Europarl}} \\ 
\hspace{-5px}Corpus & \multicolumn{2}{c}{deu$\rightarrow$spa} & \multicolumn{2}{c|}{spa$\rightarrow$deu} & \multicolumn{2}{c}{deu$\rightarrow$spa} & \multicolumn{2}{c}{spa$\rightarrow$deu}  \\
\hline
\hspace{-5px}Metrics & Acc & P@5 & Acc & P@5 & Acc & P@5 & Acc & P@5 \\
\hline
{\hspace{-5px}\scriptsize(Baseline A)} & 8.2 & 11.5 & 9.0 & 15.4 & 7.5 & 16.4 & 8.3 & 16.1 \\
{\hspace{-5px}\scriptsize(Baseline B)} & 12.3 & 26.1 & 13.5 & 27.2 & 8.1 & 14.9 & 9.5 & 19.5 \\
{\hspace{-5px}\scriptsize(Baseline C)} & 20.8 & 39.1 & 17.4 & 37.2 & 21.1 & 34.2 & 20.9 & 35.4 \\
{\hspace{-5px}\scriptsize(Baseline D)} & 15.1 & 27.5 & 15.6 & 27.9 & 20.9 & 31.6 & 21.4 & 33.1 \\
{\hspace{-5px}\scriptsize(Baseline E)} & 19.2 & 29.4 & 18.7 & 30.4 & 19.3 & 34.6 & 20.1 & 32.4 \\
{\hspace{-5px}\scriptsize seq2seq NMT} & 5.2 & 6.9 & 5.6 & 7.0 & 4.1 & 5.9 & 5.0 & 6.4\\
{\hspace{-5px}\scriptsize fairseq NMT} & 13.3 & 26.2 & 15.1 & 29.0 & 22.8 & 31.8 & 23.4 & 27.9 \\
\hspace{-5px}\ourmethod\ & \textbf{27.3} & \textbf{49.2} & \textbf{26.6} & \textbf{55.4} & \textbf{30.8} & \textbf{45.3} & \textbf{28.6} & \textbf{39.8}  \\
\hline
\end{tabular*}
\caption{Results of multilingual sentence embedding mapping experiment in terms of accuracy \%) and precision@5 (\%). The best-performing method is highlighted in bold. (Baseline A: \protect\newcite{Mikolov2013ExploitingSA}, Baseline B: \protect\newcite{dinu2014improving},  Baseline C: \protect\newcite{Smith2017OfflineBW}, Baseline D: \protect\newcite{lample2018word}, Baseline E: \protect\newcite{Schwenk2017LearningJM})}
\label{tab:multi_exp}
\end{table}

\subsection{Mapping of Low-Resource Languages}
In order to better evaluate the effectiveness and robustness of our model for diverse languages, we conduct additional experiments on low-resource languages. For comparison, we consider the state-of-the-art massively multilingual neural MT model LASER \footnote{\url{https://github.com/facebookresearch/LASER}} \cite{ArtetxeSchwenk2018}. We evaluate the mapping of low-resource languages with English on the test sets provided by them. However, as they only provide test sets but not training sets, our model is trained on comparably sized training data obtained via random sampling from Tatoeba, OpenSubtitles2018, Global Voices.\footnote{Available from \url{http://opus.nlpl.eu/}. For Irish, we incorporate some additional training data from the EUbookshop dataset} 
We adopt equivalent preprocessing steps such as filtering certain special characters and eliminating duplicate pairs. 

The results in Table \ref{tab:lowresource} confirm that simple word averages can be aligned with a broadly similar level of accuracy. This is obtained although our method does not have access to word order information and is not trained on the rich massively multilingual data used to train LASER, but only on the respective single language pairs.

\subsection{Zero-Shot Multilingual Mapping}
We also evaluate multilingual training, which entails mapping two source languages (Spanish and German) both to the same target language space (English) without any parallel data connecting the two source languages.

From both Tatoeba and Europarl, we each take 44,280 Spanish $\longleftrightarrow$ English sentence pairs and 44,280 German $\longleftrightarrow$ English sentence pairs. 
To train the baselines, we also collect the same number of German $\longleftrightarrow$ Spanish pairs for which the bilingual baselines make use of dedicated supervision, while our method does not receive any pairings at all for this language pair. Additionally, our proposed method only utilizes 20\% of parallel training data and an equal amount of non-parallel data for training, while all the  baselines take 100\% training data with parallel labels. During the training process, we alternate over mini-batches with Spanish--English pairings and German--English pairings. The number of test queries is 490 for all experiments. 

The results are reported in Table \ref{tab:multi_exp}. Though seq-to-seq models can learn full-fledged neural translation models, they do not fare particularly well in resource-constrained scenarios with limited training data. Particularly, even with a parallel sentence pair percentage of just 20\%, our model outperforms many baselines that utilize the total amount of training data. Moreover, retrieval accuracies between two source languages German and Spanish obtained by our model are very competitive with baselines receiving supervision for that language pair. Note that in our method, we do not provide any direct pairwise mapping. This proves the effectiveness of our zero-shot Multilingual mapping.

\section{Conclusion}

Our study shows that despite their simplicity, word vector averages can serve as reasonably strong cross-lingually projectable sentence representations. To this end, we have presented the \ourmethod\ model to align  such representations via an adversarial approach that requires only small amounts of parallel data. 
We obtain competitive results, although our method does 
not obtain any information about the word order in the input sentences.
Our results in a series of retrieval experiments on both short 
and long
sentences outperform previous work by a substantial margin.

{\small
\fontsize{10pt}{10pt} \selectfont
\bibliographystyle{aaai}
\bibliography{AAAI-FuZ.2038}
}

\end{document}